\pdfoutput=1

\documentclass[11pt]{article}

\usepackage[final]{acl}
\usepackage{times}
\usepackage{latexsym}

\usepackage[T1]{fontenc}

\usepackage[utf8]{inputenc}

\usepackage{microtype}

\usepackage{inconsolata}

\usepackage{graphicx}

\usepackage{tablefootnote}
\usepackage{amsmath}
\usepackage{subcaption}
\usepackage{graphicx}
\usepackage{xcolor}
\usepackage{multirow}

\usepackage{diagbox} 
\usepackage{booktabs}
\usepackage[colorinlistoftodos]{todonotes}
\title{The Role of Prosody in Spoken Question Answering}

\author{Jie Chi \thanks{This work was done during an internship at Apple MLR.} \\
University of Edinburgh\\
  \texttt{jie.chi@ed.ac.uk}  \\\And
  Maureen de Seyssel\and Natalie Schluter\\
  Apple MLR\\
  \texttt{\{mdeseyssel,natschluter\}@apple.com}
  }

\begin{document}
\maketitle
\begin{abstract}
Spoken language understanding research to date has generally carried a heavy text perspective. Most datasets are derived from text, which is then subsequently synthesized into speech, and most models typically rely on automatic transcriptions of speech. This is to the detriment of prosody--additional information carried by the speech signal beyond the phonetics of the words themselves and difficult to recover from text alone. In this work, we investigate the role of prosody in Spoken Question Answering. By isolating prosodic and lexical information on the SLUE-SQA-5 dataset, which consists of natural speech, we demonstrate that models trained on prosodic information alone can perform reasonably well by utilizing prosodic cues. However, we find that when lexical information is available, models tend to predominantly rely on it. Our findings suggest that while prosodic cues provide valuable supplementary information, more effective integration methods are required to ensure prosody contributes more significantly alongside lexical features.

\end{abstract}

\section{Introduction}
\label{sec:intro}
Prosody, which is characterized by elements of speech beyond orthographic words, such as pitch, stress and rhythm, plays a critical role in both speech production and perception. It has been shown to impact how people perceive speech, with difficulties often arising when the natural variability in prosodic structure is limited, as is the case with synthetic speech \citep{winters2004perception, wester16_speechprosody}.  In human listening comprehension,  prosodic cues are essential in guiding listeners through the process of interpreting spoken language \cite{Buck_2001,keskin2019role}. The incorrectly-stressed elements in speech can also cause listeners to make incorrect inferences \cite{field2005intelligibility}. Motivated by these linguistic findings, researchers have explored how prosody can be leveraged in speech-related tasks computationally. One of the primary tasks in this area is Spoken Language Understanding (SLU), which focuses on extracting meaningful information from spoken language input. Unlike Natural Language Understanding (NLU), which primarily deals with text-based information, SLU incorporates the added complexity of processing signal made of prosodic features such as intonation, stress, and pauses.  In this work, we use the term \textit{lexical} information to refer to information that is also present in the text in its orthographic form. That is, info that encompasses phonetic information from the speech signal. Therefore, in our work, we categorize anything that is not lexical as \textit{prosodic} information.

\begin{figure}
    \centering
    \includegraphics[width=\linewidth]{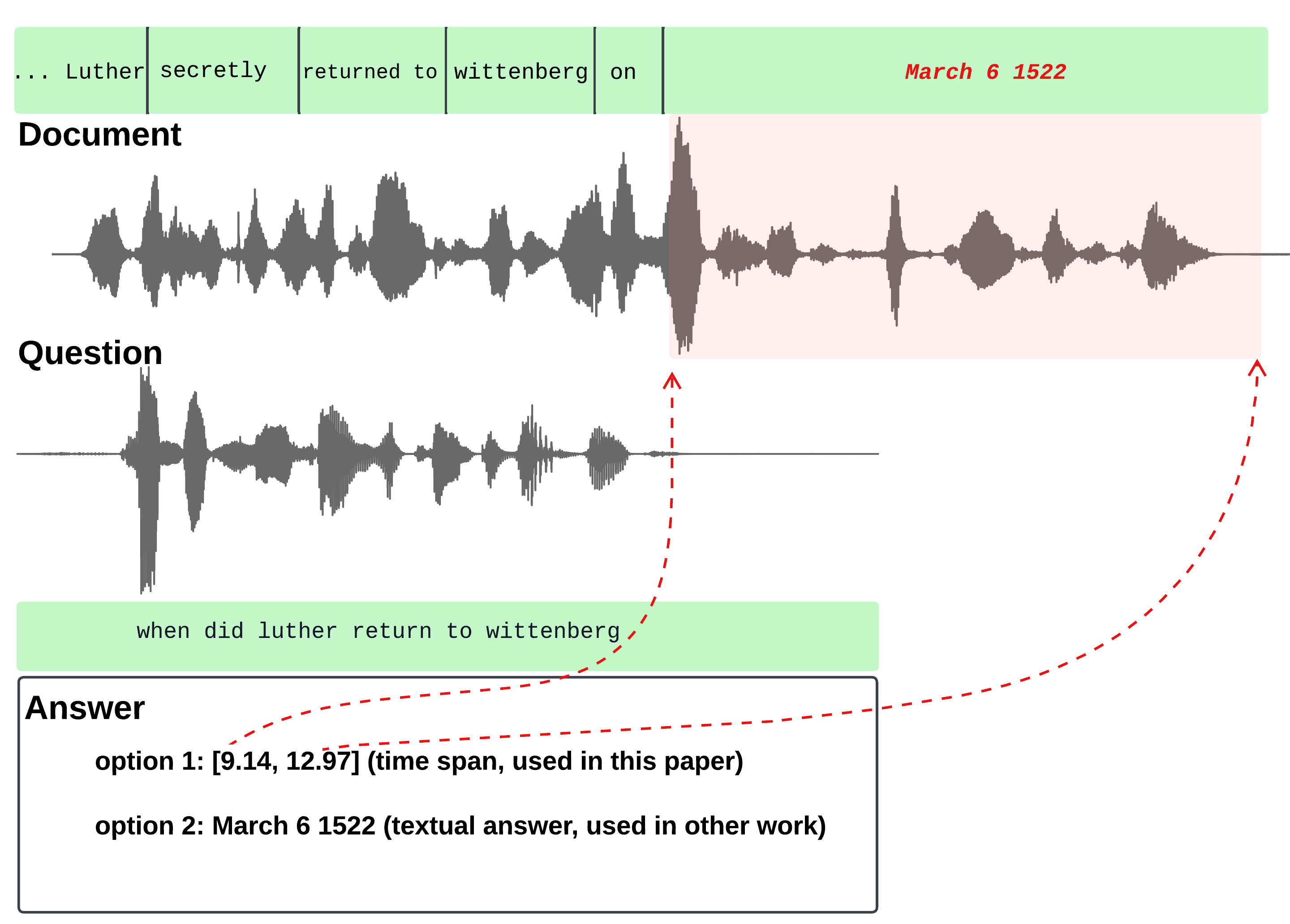}
    \caption{Illustration of the SQA format}
    \label{fig:sqa_example}
\end{figure}

Recent advances in NLU research have significantly impacted SLU through the use of a cascade approach. This approach consists of two key components: an Automatic Speech Recognition (ASR) model that transcribes speech into text, followed by an NLU model that is fine-tuned for specific downstream tasks. 
Spoken Question Answering (SQA) is one of the challenging SLU tasks, and takes the form of listening comprehension. In SQA, the input consists of a spoken passage accompanied by a question about that passage, and the model is required to provide the correct answer. This answer can take the form of timestamps in the passage, as used in this paper and previous works \cite{spokensquad,lin22c_interspeech} , or it can be a textual output \cite{shon2024discreteslulargelanguagemodel}. Figure \ref{fig:sqa_example} illustrates the typical input-output structure of an SQA task. The evolution of SQA research has gradually shifted from synthetic speech datasets to those based on natural speech. In early studies, researchers utilized Text-To-Speech (TTS) systems to convert existing Textual Question Answering (TQA) datasets into large-scale SQA corpora \cite{spokensquad, lin22c_interspeech, unlu-menevse-etal-2022-framework}. However, since SQA means prosodic information is available in the input, the prosodic characteristics of synthetic speech may not accurately represent those found in natural speech \cite{wester16_speechprosody, Clark2019EvaluatingLT, chan24_speechprosody}. This has led to concerns about the effectiveness of using synthetic speech for tasks where prosody plays a crucial role. To address these limitations, recent efforts have focused on integrating more natural speech into SQA datasets. Some studies have developed small test sets read by human speakers \cite{lin22c_interspeech}, while others have explored hybrid datasets where the questions are recorded by humans, but the passages are synthetically generated \cite{wu2024heysquadspokenquestionanswering}. Additionally, there has been work on creating training sets by sourcing spoken documents relevant to each question from external natural speech corpora \cite{shon-etal-2023-slue}.

Given the availability of the natural SQA training datasets, we aim to explore whether models can utilize prosody when comprehending speech, as humans do. The cascade approach, however, is not well-suited for this task, as prosodic information is typically lost after the transcription stage, and recovering prosody from text has been shown to be difficult \cite{talman-etal-2019-predicting}. Although some research has attempted to explicitly incorporate word-level prosodic features into NLU models \cite{tran-etal-2018-parsing, Tran2019OnTR}, the errors from ASR and alignments tend to propagate, resulting in ill-formed inputs and significantly impacting performance. Recent developments, particularly in Self-Supervised Learning (SSL) representations, have enabled researchers to bypass explicit transcription through end-to-end models \cite{chuang20b_interspeech} or by using discrete units as pseudo-text \cite{lin22c_interspeech}, which latter is the framework adopted in this paper.

In this work we are motivated to gain a comprehensive understanding of how prosodic information, distinct from the lexical information used in NLU tasks, contributes to the SQA task. Specifically, two key research questions are investigated: 1) \textit{Is prosodic information sufficient for SQA tasks?
as intonation, pitch, and pauses, signal important structural and emphatic aspects of speech}, and 2) \textit{Do SQA models utilize prosodic information when lexical information is also present?}

To address these questions, we carefully design two experimental conditions of our dataset: one that approximates prosodic information only, and another that approximates lexical information only. Directly disentangling prosodic and lexical content in speech is a complex challenge that remains unsolved \cite{quamer2024disentangling, skerry2018towards}. Among the various approaches proposed to achieve delexicalization and isolate the contribution of prosody, applying a low-pass filter has been one of the most widely used techniques, in psycholinguistics but also in modelling \cite{goldman,0fea2477791549a785cbdd7e0c058ca5,audibert2023evaluation}; it can preserve an approximation of the prosodic features while removing most of the discriminating information that phonetically delineates orthographic words \cite{mehler1988precursor}. 

Hence, for the prosodic condition, we apply a low-pass filter to remove information above a certain cutoff frequency, ensuring that most lexical information is excluded from the speech signal. In contrast, for the lexical condition, we flatten both pitch and intensity to eliminate most prosodic variation. While prosodic and lexical information are not completely disentangled, our experiments show that the reduction of these elements is sufficient to prevent significant interference of one variable over the other in the results.

Through controlled experiments, we demonstrate that prosodic information alone can, to some extent, guide models to answer questions in SQA tasks. However, while prosody offers meaningful complementary cues, we find that models predominantly rely on lexical information when it is available. By providing a deeper understanding of prosody’s role in SQA, we hope to pave the way for future work on developing more robust models capable of leveraging both lexical and prosodic information effectively, particularly in situations where lexical information is limited or degraded.

\section{Related work}
To date, SLU research has mostly involved first identifying word sequences.  As such, there has been a focus on integrating prosodic information into ASR models. Previous research has used prosody by conditioning the acoustic and pronunciation modelling on prosodic features \cite{Shriberg, 1561280}, simultaneously predicting prosodic events \cite{chen2003prosody,Hasegawa-Johnson}, or incorporating prosodic information in N-best rescoring in hybrid ASR systems \cite{4218240,Ananthakrishnan, huang10_speechprosody}. However, current state-of-the-art ASR models do not model prosody explicitly. 

Researchers have also explored using prosody to help other tasks.  Assuming known time alignments, incorporating word-level prosodic features has yielded improvements in constituent parsing of conversational speech \cite{tran-etal-2018-parsing, Tran2019OnTR}. Prosody has also been shown helpful in topic tracking \cite{guinaudeau11_interspeech}, dialogue act classification \cite{wei2022neuralprosodyencoderendroend} speech to intent \cite{rajaa23_interspeech}, and emotion recognition \cite{Luengo2005AutomaticER,naderi2023cross}. These studies modeled prosodic patterns either at the word or utterance level by averaging frame-level features such as pitch and intensity, or by using other hand-selected prosodic features. With the rise in popularity of neural networks, prosodic patterns can now be more effectively captured and modeled directly through CNNs, allowing for a more comprehensive representation of the prosodic features without the need for manual selection. 

However, with the emergence of SSL models \cite{Baevski2020wav2vec2A, hsu2021hubert, chen2022wavlm}, prosody is usually not explicitly modeled as a separate feature. Instead, it has been shown that these models capture prosodic information implicitly within their learned representations, alongside other linguistic features. SSL models are pre-trained on large amounts of unlabeled audio data, learning representations by predicting missing portions of the input signal or clustered latent speech units. There has been extensive research exploring the utility of SSL representations in prosody-related tasks, and it has been concluded that these representations encode prosodic information such as gender and speaker identity \cite{deseyssel22_interspeech, oli, Mukhtar}. These representations have also been successfully applied to tasks such as emotion recognition, speaker identification, and intonation analysis \cite{10023234}. To leverage the capabilities of advanced language models, k-means clustering or other quantization approaches are typically used in conjunction with SSL representations to reduce the length of the input sequences. It has been observed that even within these discrete units, prosodic information is preserved, resulting in relatively low error rates for speaker and gender classification, particularly when more clusters are used \cite{deseyssel22_interspeech}. This suggests that these discrete units retain key prosodic cues, which we use to represent speech in our study, allowing us to investigate the role of prosody in SQA tasks more effectively.

\section{Methods}
To investigate the role of prosody in SQA, in addition to the original datasets which combine both lexical and prosodic information (i.e., the dataset in its \emph{natural condition}), we also designed a set of experiments that systematically examine the effects of prosodic information and lexical information individually. Our approach involves three main stages: data preparation, model training, and evaluation.
\subsection{Data preparation}
\subsubsection*{SLUE-SQA-5 dataset}

We use the SLUE-SQA-5 dataset \cite{shon-etal-2023-slue} for this work. Unlike earlier datasets, which often rely on synthetic speech generated from TTS systems, it features naturally occurring spoken data, allowing for the study of prosody in a more realistic context. The corpus is derived from five existing TQA datasets and the questions are collected from crowd-source workers. The documents are collected by retrieving relevant documents to each question from the Spoken Wikipedia dataset \cite{KHN16.518}. All audios are from natural speech, which in this paper, we refer to as the \textit{natural condition} (as opposed to the \emph{lexical} or \emph{prosodic} conditions that we define shortly). Table \ref{tab:slue} illustrates the statistics of the corpus. In addition to train, test, and dev sets, the dataset includes a verified test set consisting of hand-picked question-document pairs from the test set, in which the document provides sufficient clues to answer the question.

\begin{table}
    \centering
     \resizebox{\linewidth}{!}{
    \begin{tabular}{c|c|c|c}
    \toprule
        dataset& questions & documents & duration (hrs) \\
         \midrule
         train & 46186&15148&244 \\
         dev & 1939&1624&21.2 \\
        test &2382&1969&25.8\\
        verified test & 408&322&4.2 \\
    \bottomrule
    \end{tabular}}
    \caption{Statistics over the SLUE-SQA-5 dataset. }
    \label{tab:slue}
\end{table}

\subsubsection*{Dataset modification}
\begin{figure}[htb!]
      \centering
	   \begin{subfigure}{\linewidth}
		\includegraphics[width=\linewidth]{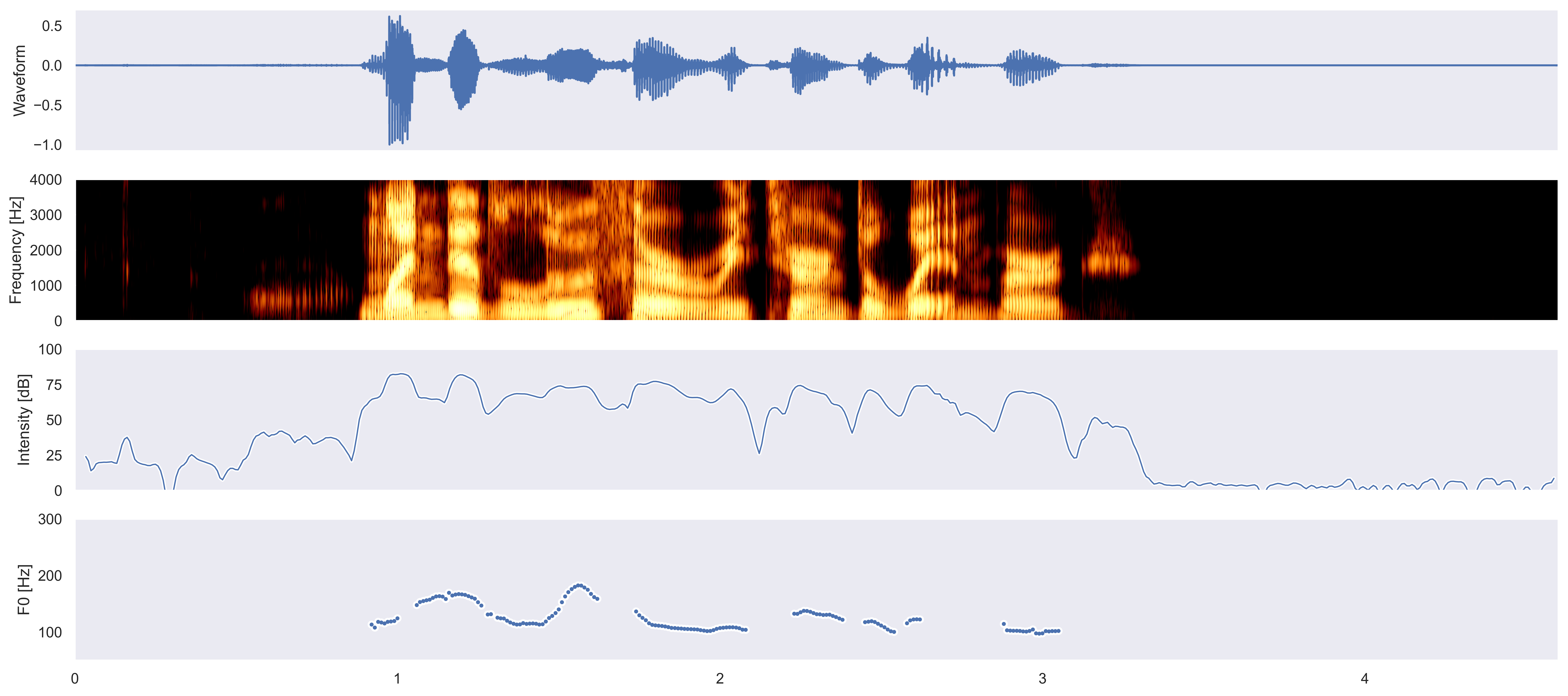}
		\caption{Natural condition}
		\label{fig:natural}
	   \end{subfigure}
	   \begin{subfigure}{\linewidth}
		\includegraphics[width=\linewidth]{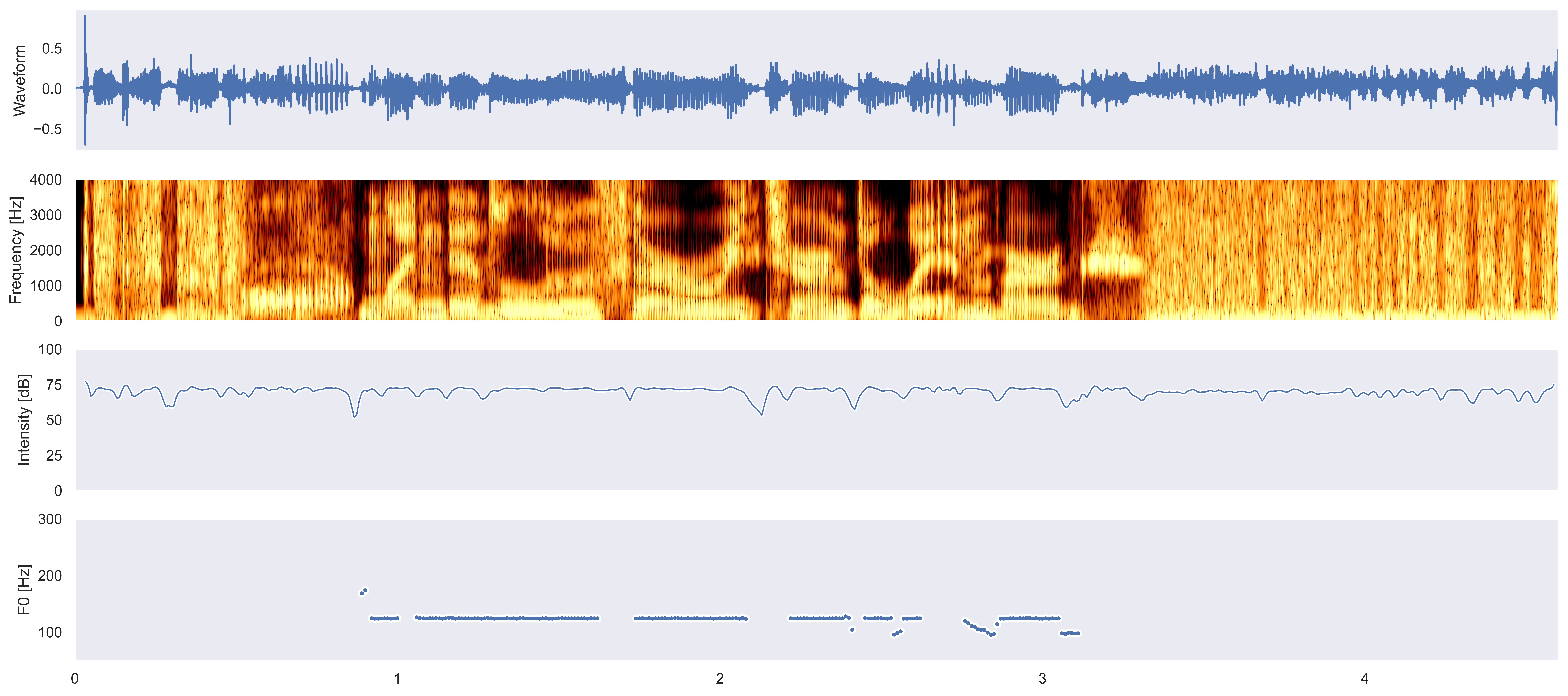}
		\caption{Lexical condition}
		\label{fig:lex}
	    \end{subfigure}
	     \begin{subfigure}{\linewidth}
		 \includegraphics[width=\linewidth]{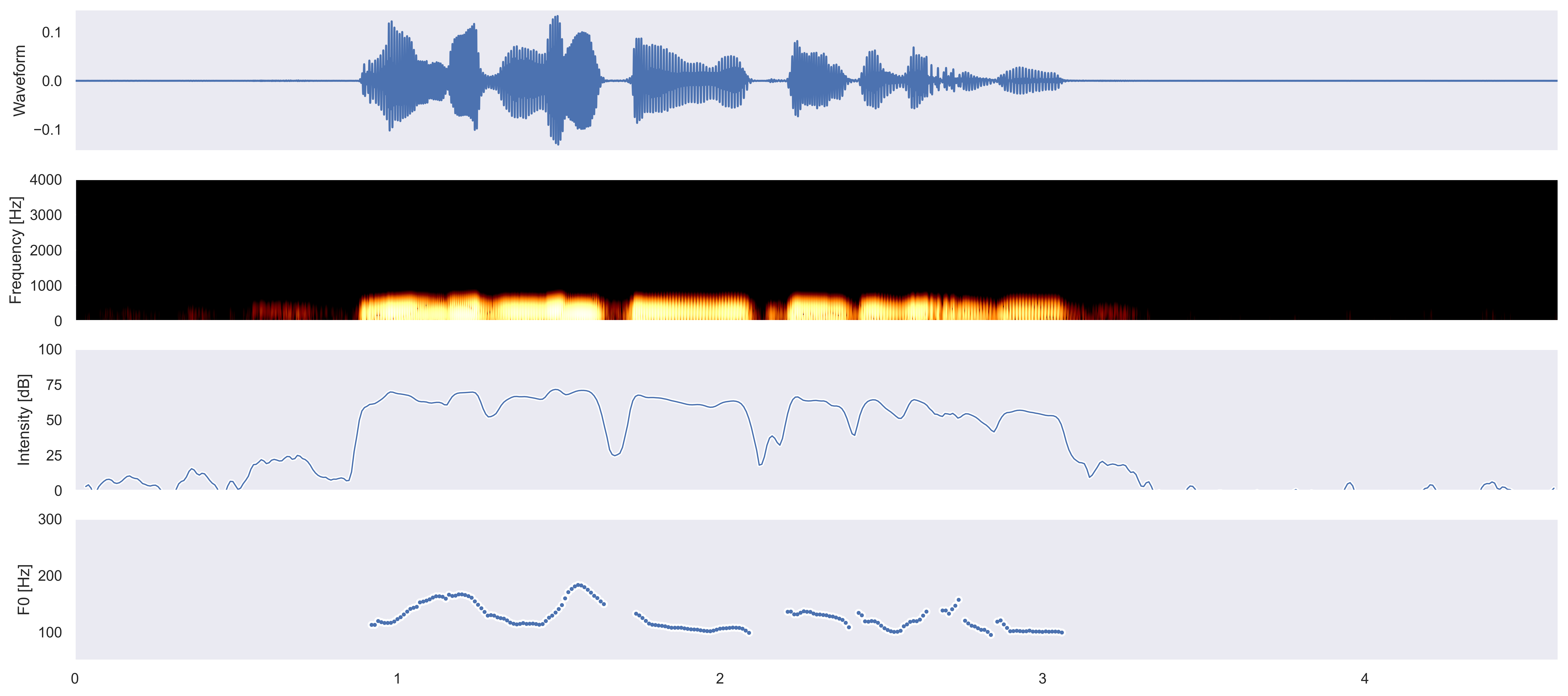}
		 \caption{Prosodic condition}
		 \label{fig:pros}
	      \end{subfigure}
	\caption{Spectrogram of the example speech under different conditions. In each sub-figure, the top plot is the waveform, the second plot is the spectrogram, the third plot is the intensity, and the bottom plot is the F0.}
	\label{fig:spectrogram}
\end{figure}

We modify the audio in two different ways with Parselmouth \cite{parselmouth, praat}, and examples of spectrograms \footnote{The corresponding audio files are provided in the supplementary materials.} for a same audio under different conditions are shown in Figure \ref{fig:spectrogram}.

In the first setting, we remove the variations in both pitch and intensity, which we refer to as the \emph{lexical condition}. This modification ensures that primarily lexical information remains, while that two of the main prosodic features,  intonation and stress are considerably reduced. This results in a non-expressive, almost robotic-like quality to the sound. Specifically, we flatten both the fundamental frequency and intensity to the average value of each utterance. This approach helps us examine how models respond when only lexical information is present, with minimal prosodic variation. In Figure \ref{fig:lex}, we can observe that the F0 and intensity contour is nearly flat. It should be noted that it is hard not to introduce the artifacts when flattening the intensity, for example, breath can become very loud and the gain inside smaller segments of silence is prominent \cite{ekstedt-skantze-2022-much}. It should still be noted that rhythm (i.e duration) is not modified here.

In the second setting which we refer to as the \emph{prosodic condition}, we apply a low-pass filter to the audio, removing high-frequency components, as shown in Figure \ref{fig:pros} where the spectrogram shows a clear cut-off above the threshold. Filtering audio by frequency inevitably affects prosodic information as well, since prosody is embedded in various frequency bands. To mitigate this, we set the cut-off frequency to 300Hz, aiming to preserve as much prosodic information as possible while reducing high-frequency lexical cues. The choice of 300Hz is based on the distribution of speech energy: vowel sounds generally lie in the range of 250 to 2000Hz, voiced consonants between 250 and 4000Hz, and unvoiced or voiceless consonants primarily occupy the 2000 to 8000Hz range \cite{cutoff}. By targeting the lower frequencies, we attempt to retain certain prosodic elements like pitch contour and rhythm, while removing higher-frequency lexical content. Section \ref{sect:rq1} explores the effects of applying different cut-off frequencies, allowing us to investigate how varying amounts of high-frequency information influence model performance. 

It is important to note that we do not aim to fully disentangle prosodic and lexical information as this would be an extremely complex task given their intertwined nature in natural speech. Instead, our objective is to generate modified versions of the dataset that either keep lexical information by suppressing prosody or reduce lexical content while retaining some prosodic cues.   We are not claiming that the remaining prosodic or lexical information is identical to its representation in natural speech, rather this research is carried out under the awareness of this is a limitation. These manipulations serve as controlled approximations, allowing us to systematically investigate the role of prosody and lexical information in SQA tasks.

\subsection{Model training}

We use the Discrete Spoken Unit Adaptive Learning (DUAL) framework \cite{lin22c_interspeech}, which consists of two main components: a Speech Content Encoder (SCE) and a Pre-trained Language Model (PLM). Unlike conventional cascade models, DUAL bypasses the reliance on ASR transcripts, and thus also the associated ASR error propagation. The SCE leverages WavLM, a self-supervised pre-trained model known for strong performance in prosody-related tasks \citep{10023234}, to encode representations directly from raw audio waveforms. These representations are then processed using k-means clustering, converting them into discrete units that are deduplicated before being fed into the PLM. 

Although deduplication could potentially discard duration information, which is an important aspect of prosody, the impact in our case is minimal. Using 1000 clusters, we observe very few repetitions, with only 8\% of units showing more than three consecutive repetitions in the verified testset, which are likely due to silence. Furthermore, higher cluster counts have been shown to retain more prosodic information, as demonstrated by performance improvements in tasks like gender and speaker classification \cite{Sicherman_2023}. 

For our SCE, we use a pretrained SpeechBrain model \cite{SB2021}\footnote{
We use the pre-trained WavLM representations available at \url{https://huggingface.co/speechbrain/SSL_Quantization/tree/main/LibriSpeech960/wavlm/LibriSpeech_wavlm_k1000_L23.pt} for our experiments. While it is possible to select representations from different layers, a thorough layer-wise analysis falls beyond the scope of this work.}, which is a WavLM Large model pretrained on the Librispeech 960-hour corpus. The representations from this model remain frozen for all our experiments. The PLM is responsible for predicting the answer span within the context passage by identifying the start and end positions, similar to a typical TQA model. Consistent with the DUAL paper, we use the Longformer-base model\footnote{\url{https://huggingface.co/allenai/longformer-base-4096}} as the PLM, a BERT-like model for long documents, pretrained on unlabeled long text documents \cite{beltagy2020longformerlongdocumenttransformer}.

For reproducibility, all configurations are detailed here. We utilize eight A100-80 GPUs with a total batch size of 128, training the models for up to 18 epochs. Following the original DUAL framework, the learning rate is warmed up over the first 500 steps. We conduct a learning rate search within the range of $[5e-6, 1e-5, 5e-5, 1e-4]$, starting with the natural condition. Once the optimal learning rate is identified, we evaluate the performance on other conditions to ensure that the selected rate is not biased toward the natural condition. Ultimately, the learning rate is fixed at $1e-5$ for all experiments. The model needs 160 GPU hrs when tracking all evaluations in Section \ref{sect:rq2}.  To account for variability, all models are trained using three different random seeds, and we report the mean and standard deviation across all results.

\begin{table*}[ht]
\centering
\resizebox{0.9\textwidth}{!}{
    \begin{tabular}{c|c|c|c|c|c|c}
        \toprule
     &\multicolumn{2}{c|}{\textbf{Natural}} & \multicolumn{2}{c|}{\textbf{Lexical}} & \multicolumn{2}{c}{\textbf{Prosodic}} \\
    &   \textbf{Test} & \textbf{Verified-Test} & \textbf{Test} & \textbf{Verified-Test} & \textbf{Test} & \textbf{Verified-Test} \\ 
        \midrule
     & \multicolumn{6}{c}{\textbf{FF1}}\\
      Natural&  \textbf{33.27 ± 1.20} & \textbf{36.01 ± 0.86}  & 26.12 ± 0.59 & 30.01 ± 1.23 & 10.04 ± 0.47 & 10.57 ± 0.87\\
      Lexical &   26.72 ± 0.41 & 29.31 ± 0.71 & \textbf{32.37 ± 0.12} &  \textbf{33.42 ± 0.55}  & 8.94 ± 0.61 & 8.19 ± 0.66 \\
      Prosodic &  8.33 ± 0.51 & 9.90 ± 2.19 & 6.96 ± 1.25 & 9.01 ± 1.07  & \textbf{18.49 ± 0.67} & \textbf{18.29 ± 1.14} \\

              \midrule
             & \multicolumn{6}{c}{\textbf{AOS}}\\
      Natural&  \textbf{29.67 ± 1.24} & \textbf{31.80 ± 0.78} & 22.47 ± 0.68 & 26.04 ± 0.97 &  6.36 ± 0.41 &  6.75 ± 0.82 \\
      Lexical &  23.13 ± 0.26 & 26.04 ± 0.97 & \textbf{28.63 ± 0.02} & \textbf{29.30 ± 0.68} & 5.95 ± 0.46 & 5.57 ± 0.32\\
      Prosodic &  5.39 ± 0.39 & 6.79 ± 1.65 & 4.37 ± 0.94 & 5.99 ± 0.94 & \textbf{13.97 ± 0.73} & \textbf{13.86 ± 0.98} \\
    \bottomrule
    \end{tabular}}
\caption{Results for different training and testing conditions (natural, lexical, and prosodic) on the test and verified test set. Bold diagonal cells indicate results when training and tesing conditions are the same. }
\label{tab:individual result}
\end{table*}

\begin{figure}
    \centering
    \includegraphics[width=\linewidth]{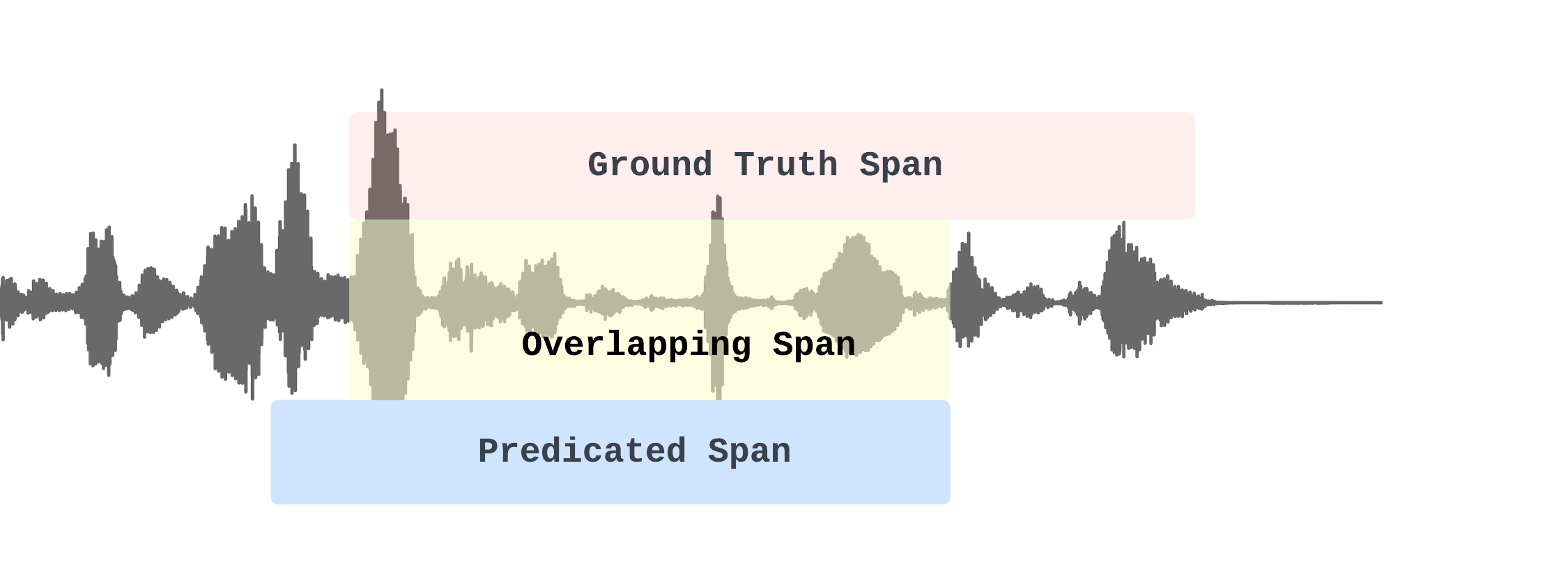}
    \caption{Illustration of ground truth span, predicted span and overlapping span for evaluation.}
    \label{fig:overlap}
\end{figure}

\subsection{Evaluation}
To evaluate the performance of our models, we employ two key metrics: Frame-level F1 (FF1) score \cite{chuang20b_interspeech}  and Audio Overlapping Score (AOS) \cite{spokensquad}, both of which are commonly used in SQA tasks to assess model accuracy with respect to time-based predictions.

\textbf{FF1 score} is an adaptation of the standard F1 score used in text-based question answering (TQA) tasks. While in TQA, the F1 score is calculated based on token-level matches between predicted and ground-truth answers, in SQA, the answers are temporal segments of audio rather than discrete tokens. Therefore, FF1 measures the precision and recall of frame-level matches between the predicted and actual answer spans, as shown in Figure \ref{fig:overlap}. The equation is as follows

\begin{equation*} 
\begin{split}
Precision & = \frac{\textit{\text{Overlapping Span}}}{\textit{\text{Predicted Span}} }\\
Recall  & = \frac{\textit{\text{Overlapping Span}}}{\textit{\text{Ground Truth Span}}}\\
FF1  & = \frac{2 \times \textit{\text{Precision}} \times \textit{\text{Recall}}}{\textit{\text{Precision}} + \textit{\text{Recall}}}  \\
\end{split}
\end{equation*}

\textbf{AOS} on the other hand, provides an additional evaluation by measuring the overlap between the predicted and ground-truth answer spans using the intersection-over-union  ratio at the frame level. The equation can be written as follows
\begin{equation*} 
\begin{split}
AOS  & = \frac{\textit{\text{Overlapping Span}}}{\textit{\text{Predicted Span}} \cup \textit{\text{Ground Truth Span}}}  \\
\end{split}
\end{equation*}

\section{Experiments and results}
\label{sec:exp}
To explore the role of prosodic information in SQA tasks, we design two stages of experiments to answer the research questions stated in Section \ref{sec:intro}.

\subsection{Is prosodic information sufficient for SQA tasks?}
\label{sect:rq1}
First, we train the model separately over our three different data conditions: \textit{natural}, \textit{lexical} and \textit{prosodic} and test on the corresponding conditions. The results are shown in Table \ref{tab:individual result}. We also establish a \textit{chance-level} baseline by generating white noise speech using a normal distribution with the same duration as the documents in the verified-test set in order to simulate the model's performance in the absence of all meaningful information. Evaluating all models across different seeds for generating the white noise speech, the best FF1 and AOS  are $6.03 \pm 0.16$ and $3.29 \pm 0.09$, respectively. This serves as the default result when random predictions are made, providing a baseline for comparison with the model's performance on actual data. 

As expected, when the training and testing conditions are the same, the model performs best under the natural condition, followed by the lexical condition. But interestingly, the prosodic condition also performs reasonably well, far better than the chance level baseline \footnote{The DUAL baseline, using \textit{ wav2vec} as the encoder, achieved an FF1 score of $23.1$ on the same natural condition for verified-test set, as reported in \cite{shon-etal-2023-slue}. We include their result here not for direct comparison of the models, but to highlight that the results obtained using prosodic information in our experiments are very close to a functional baseline using both prosodic and lexical information, significantly better than random performance.}. 

The model performs the worst when trained and tested on some combination of lexical and prosodic conditions, as this combination has the least information overlap among all the tested configurations. Although the results for natural and lexical conditions are not identical, the relatively close performance suggests that the model relies heavily on lexical information, as both conditions include it. When trained on either natural or lexical condition, the model performs well on the other condition, indicating that the model can generalize effectively between natural and lexical information. However, when tested on the prosodic condition, performance drops significantly. This pattern is mirrored when the model is trained on prosodic information and tested on natural or lexical conditions. These results highlight that, although prosody alone provides meaningful information for SQA from the results, it cannot fully compensate for the absence of lexical content. This ability to generalize only occurs despite a domain mismatch in prosody between lexical and natural conditions, suggesting that the model prioritizes lexical cues over prosodic variations when both are available.

\begin{figure}
    \centering
    \includegraphics[width=\linewidth]{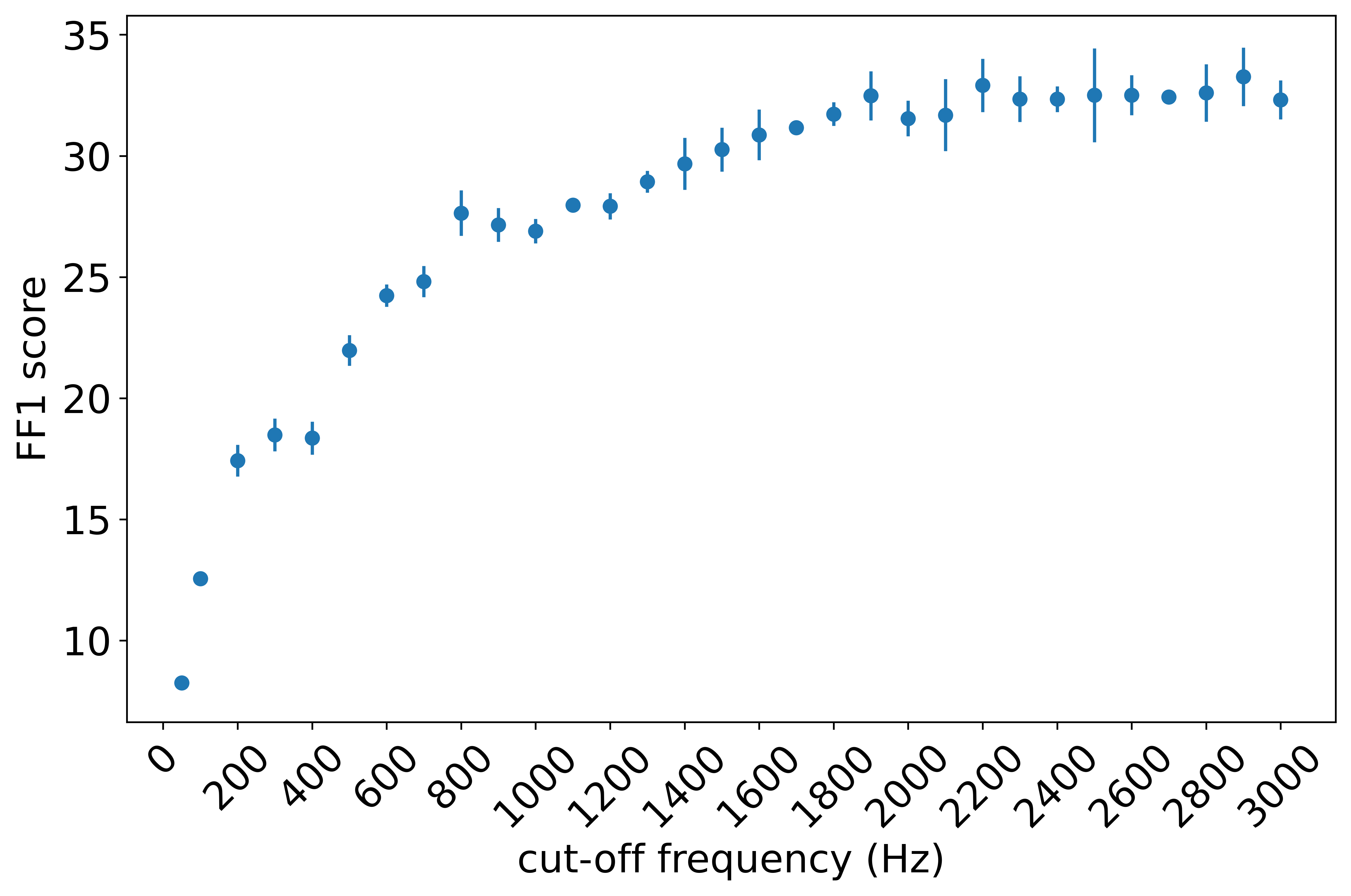}
    \caption{Performance on the test set with different cut-off frequencies}
    \label{fig:cutoff}
\end{figure}

\paragraph{Can the prosodic condition scores stem from leftover lexical information?} One possible concern is whether the performance observed in the prosodic condition could be purely influenced by \textit{residual lexical information}. To address this, we select a cut-off frequency of 300Hz, which theoretically removes the majority of energy associated with both vowels and consonants, given that vowel sounds typically lie in the range of 250–2000Hz, and consonants span from 250Hz to as high as 8000Hz. To further investigate, we conduct additional experiments using different cut-off frequencies (from 50 to 3000Hz) to assess the impact of filtering on model performance. The result is presented in Figure \ref{fig:cutoff}. We observe there is no significant performance drop when the cut-off frequency is above 1800Hz and between the 200Hz and 400Hz. This suggests that frequencies above 1800Hz are primarily associated with high-frequency sounds such as fricatives and certain consonants, which are less critical for understanding the core content of speech. However, the model's performance gradually decreases as the cut-off frequency is lowered from 1800Hz to 400Hz, indicating that most lexical information is included within this range as it contains many formant frequencies of vowels and essential cues for consonants. 
The sharp performance drop below 200Hz can be attributed to the loss of crucial prosodic information, especially F0, which plays a key role in intonation, and stress patterns. Therefore, a cut-off frequency between 200Hz and 400Hz offers a good compromise, retaining enough prosodic information while effectively removing most lexical content. 

To bring further support to our observations, we present the WER results for the WavLM-CTC model, which was trained on 960 hours of Librispeech data and evaluated on both the SLUE-SQA-5 test and verified test datasets. As shown in Table \ref{tab:asr result}, the prosodic conditions in Librispeech and the test sets were matched by applying the same cutoff-frequency low-pass filter to all data, ensuring consistency. Similarly we find when frequencies drop below 200 Hz, the audio becomes completely unintelligible, even when the model is trained on data processed under the same condition. In contrast, when the cutoff frequency is above 200 Hz, the WER quickly decreases to approximately 50\%. This result is not unexpected, as speech recognition may not be the ideal task for evaluating residual lexical information; rather, recognition reflects the process by which these lexical and prosodic properties are perceived, not a direct measure of the information present. Previous work has demonstrated that prosodic cues can be beneficial for the task \cite{VICSI2010413, Bhardwaj}, and there is notable redundancy between lexical and prosodic channels that further impacts performance \cite{wolf-etal-2023-quantifying}. Nevertheless, while the exploration of prosody's impact on ASR falls outside the scope of this study, our cutoff frequency analysis confirms that SQA performance is not solely due to residual lexical content.

\begin{table}[ht]
\centering
\begin{tabular}{c|c|c}
    \toprule
    Cut-off (Hz) & Test & Verified-Test \\
    \midrule
    50   & 133.1 & 147.3 \\
    100  & 144.5 & 133.5 \\
    200  & 80.3  & 72.3  \\
    300  & 57.5  & 44.5  \\
    400  & 49.2  & 36.8  \\
    500  & 45.6  & 33.9  \\
    \bottomrule
\end{tabular}
\caption{WER results for the WavLM-CTC model (960h Librispeech) evaluated on SLUE-SQA-5 test and verified datasets. Consistent prosodic condition was ensured for training and evaluation.}
\label{tab:asr result}
\end{table}

\paragraph{Is the question relevant in the prosodic condition?} 

In the context of SQA, prosodic information, such as intonation, pitch, and pauses, signals important structural and emphatic aspects of speech. These cues can highlight portions of the context that are more likely to contain relevant information. For example, changes in intonation may signal the introduction of key points, while pauses and shifts in pitch can emphasize certain phrases or concepts. As a result, prosodic information can help narrow down the likely locations of the answer within the context, even when lexical information is absent or reduced. However, one concern is whether prosodic information alone can meaningfully contribute to SQA if it can disregard the actual \textit{question}. If prosody merely highlights key parts of the context without connecting them to the question, contribution of the question might be limited. To explore this, we conduct an experiment in which questions and contexts for the verified test set were randomly paired. In this setup, the model’s performance dropped significantly, reaching levels similar to those observed when lexical information was present but not prosodic information. This drop in performance indicates that prosodic cues alone are insufficient to fully answer the question, as they do not directly convey the question’s relationship to the relevant parts of the context.

\begin{table}[ht]
\centering
    \begin{tabular}{c|c|c}
        \toprule
            & \textbf{FF1} & \textbf{AOS} \\ 
        \midrule
       Natural & 17.09 ± 0.96  & 14.79 ± 0.89  \\
       Lexical & 17.26 ± 0.33  &  14.74 ± 0.25 \\
       Prosodic & 9.77 ± 0.54  & 7.05 ± 0.57 \\
    \bottomrule
    \end{tabular}
\caption{Results for the same training and testing conditions on the random-paired verified test set. }
\label{tab:random result}
\end{table}

Nevertheless, it is important to note that the model still performed better than random chance, suggesting that prosodic cues provide some utility even in the absence of a meaningful connection between the question and context. These cues likely highlight segments of the passage that are perceived as more important or emphasized, helping the model identify areas where relevant information might be located. This explains why the model outperforms a chance-level baseline even when the question-context alignment is disrupted. In summary, while prosody alone cannot entirely guide the model to the correct answer, it serves as a helpful supplementary signal that directs attention to key parts of the passage.

\begin{figure*}
\centering
\includegraphics[width=\textwidth]{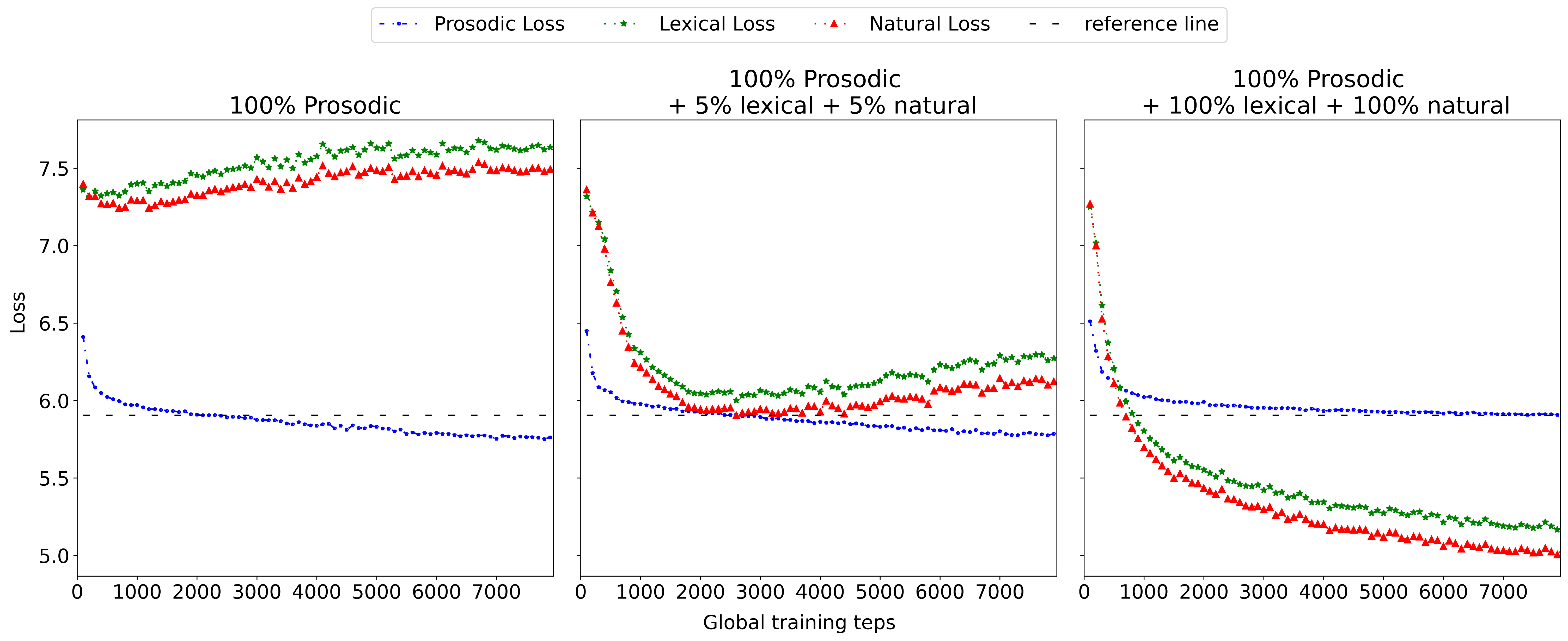}
\caption{Evaluation loss across different conditions. From left to right, the model is trained on (1) the full prosodic training set, (2) a combination of the full prosodic training set and 5\% from both the lexical and natural training sets, and (3) the full training sets for all conditions. The reference line indicates the lowest prosodic loss when the model is trained on all conditions..}
\label{fig:loss}
\end{figure*}

\subsection{Do SQA models utilize prosodic information when lexical information is also present?}
\label{sect:rq2}

From the results presented in Table \ref{tab:individual result}, we observe that the model tends to prioritize lexical information from the similar performance on both the natural and lexical sets across different configurations. To better understand this behavior, we conduct experiments for the prosodic condition when combining training with portions of 0\%, 5\%, and 100\% of the training sets from other two conditions. We then track the evaluation loss on all conditions throughout the training process. As shown in Figure \ref{fig:loss}, the evaluation loss trends clearly demonstrate that the model predominantly relies on lexical information. When this lexical information is absent during training, the model learns to utilize prosodic information as indicated by the decrease in prosodic evaluation loss. However, even when only 10\% of the training data contains lexical information and the overwhelming majority consists of prosodic data, the model quickly learns from the lexical data. This is reflected by the rapid decrease in evaluation loss for both the lexical and natural sets, which soon approach the same level as the prosodic loss. When equal amounts of data from each condition are provided, we observe a more rapid decrease in loss for both the natural and lexical sets, whereas the prosodic set exhibits higher loss than the other two training conditions with less lexical information, as indicated by the reference line.  This suggests that, when given access to both lexical and prosodic features, the model primarily uses lexical information, possibly because lexical features offer a more straightforward path to understanding and answering questions based on the content of the speech. In contrast, prosodic cues, though helpful, do not seem to be the model's primary source of information in these settings.

\section{Conclusion}
Through a series of controlled experiments, we explore the role of prosody in SQA tasks. Our findings demonstrate that while lexical information remains the dominant feature in models trained on both prosodic and lexical data, prosody still provides meaningful complementary cues. In experiments where prosodic information was isolated, the model performed reasonably well, indicating that prosody alone can guide the model toward identifying relevant segments in the context. This underscores the independent value of prosodic features such as intonation, stress, and pauses in guiding the model’s understanding of spoken language.

However, when both prosodic and lexical information were available, the model predominantly rely on lexical cues, as they offer a more direct path to understanding the meaning of the speech. Even when the amount of lexical information in the training data was reduced to just 10\%, the model continued to prioritize and learn from these features over prosodic cues. This suggests that while prosody is useful, it is often overshadowed by lexical content in tasks where both types of information are present.

Additionally, our analysis of random question-context pairings reveal that prosodic cues alone cannot fully guide the model to the correct answer without considering the relationship between the question and the relevant context. Nevertheless, the model still performed above random chance, suggesting that prosodic information highlights important parts of the context, even when it cannot provide the complete answer.

In conclusion, while prosodic information plays a valuable role in SQA, its contribution is secondary to lexical content when both are present. Future work should focus on developing models that can better integrate prosodic and lexical information to fully leverage the richness of spoken language, especially in scenarios where lexical information is degraded or limited.

\section{Limitations}

We have ensured full reproducibility of our results by using both an open-source model and original dataset, and providing detailed instructions (including hyperparameters) for replicating our experimental conditions and results.
We acknowledge limitations in our work, with the primary challenge being the difficulty in making the prosodic and lexical information fully independent when designing our conditions. Indeed, in the prosodic condition, while we tried to minimize lexical information, there remains the possibility of some residual lexical cues contributing to the model's performance. Moreover, by applying a low-pass filter, we also degrade the quality of the prosodic information, potentially artifically lowering the scores related to prosodic only information. Future work could explore more sophisticated methods of explicitly modelling prosodic features separately from lexical ones.

Another limitation relies in the choice of layer used to extract the representations. While we used one of the deeper layers of the model to extract our discrete units, as it has been suggested that is where semantic information is the strongest \cite{9688093}, it is possible that prosodic information is more heavily encoded in earlier layers. Further exploration of the different representations could bring more light on the role of prosody on SQA.

Furthermore, our study used SLUE-PHASE2, an extractive SQA dataset where answers are specific spans of audio within a passage. This approach limited our investigation to tasks requiring literal comprehension, such as identifying places, names, or dates. While this provides insight into how prosody helps in locating specific information, it would be valuable to extend this research to open-ended SQA tasks, where prosodic information may play a more significant role in guiding models to generate nuanced and contextually appropriate responses. 

Finally, future work should explore how prosody influences inferential comprehension, where emotions, thoughts, and intentions are inferred from the speech. In these tasks, prosody could offer important cues that go beyond the lexical content, enriching the model's understanding of more abstract or emotional aspects of the spoken language.

\bibliography{acl_latex}




\end{document}